\documentclass[a4paper,twoside]{article}

\usepackage{epsfig}
\usepackage{subcaption}
\usepackage{calc}
\usepackage{amssymb}
\usepackage{amstext}
\usepackage{amsmath}
\usepackage{amsthm}
\usepackage{multicol}
\usepackage{pslatex}
\usepackage{apalike}
\usepackage{algorithm2e}
\usepackage[bottom]{footmisc}

\usepackage{gensymb}
\usepackage{booktabs}
\usepackage{subcaption}
\usepackage{xcolor}
\let\bibhang\relax
\usepackage{natbib}
\usepackage{multirow,makecell}

\usepackage{SCITEPRESS}     % Please add other packages that you may need BEFORE the SCITEPRESS.sty package.

\begin{document}

\title{How Quality Affects Deep Neural Networks in Fine-Grained Image Classification}

\author{
\authorname{
Joseph Smith\sup{1},
Zheming Zuo\sup{1},
Jonathan Stonehouse\sup{2}
and Boguslaw Obara\sup{1}
}
\affiliation{\sup{1}School of Computing, Newcastle University, Newcastle upon Tyne, UK}
\affiliation{\sup{2}Procter and Gamble, Reading, UK}
\email{\{j.smith57, zheming.zuo, boguslaw.obara\}@newcastle.ac.uk, stonehouse.jr@pg.com}
}

\keywords{Image Quality Assessment, Neural Networks, Fine-Grained Image Classification, Mobile Imaging.}
% \keywords{The paper must have at least one keyword. The text must be set to 9-point font size and without the use of bold or italic font style. For more than one keyword, please use a comma as a separator. Keywords must be titlecased.}

\abstract{In this paper, we propose a No-Reference Image Quality Assessment (NRIQA) guided cut-off point selection (CPS) strategy to enhance the performance of a fine-grained classification system. Scores given by existing NRIQA methods on the same image may vary and not be as independent of natural image augmentations as expected, which weakens their connection and explainability to fine-grained image classification. Taking the three most commonly adopted image augmentation configurations -- cropping, rotating, and blurring -- as the entry point, we formulate a two-step mechanism for selecting the most discriminative subset from a given image dataset by considering both the confidence of model predictions and the density distribution of image qualities over several NRIQA methods. Concretely, the cut-off points yielded by those methods are aggregated via majority voting to inform the process of image subset selection. The efficacy and efficiency of such a mechanism have been confirmed by comparing the models being trained on high-quality images against a combination of high- and low-quality ones, with a range of 0.7\% to 4.2\% improvement on a commercial product dataset in terms of mean accuracy through four deep neural classifiers. The robustness of the mechanism has been proven by the observations that all the selected high-quality images can work jointly with 70\% low-quality images with 1.3\% of classification precision sacrificed when using ResNet34 in an ablation study.}

\onecolumn \maketitle \normalsize \setcounter{footnote}{0} \vfill

\section{Introduction}
\label{sec:intro}

% Convolutional Neural Networks (CNNs) are at the centre of many fine-grained image classification systems (\cite{razavian2014cnn}), \emph{e.g.} object detection and scene understanding in self-driving cars (\cite{gupta2021deep} and white blood cell classification in medicine (\cite{ha2022fine}), 

Convolutional Neural Networks (CNNs) are at the centre of many fine-grained image classification systems (\cite{razavian2014cnn,gupta2021deep,ha2022fine})
and have been shown to be susceptible to low-quality images, particularly blurring and noise, which can deteriorate the precision of model prediction (\cite{Dodge:2016}). This issue could be even more significant when dealing with images from an uncontrolled environment, where the qualities could vary considerably (\cite{sabbatini2021computer,aqqa2019understanding}).

Fine-grained image classification tasks are even more vulnerable to low image quality than coarse-grained tasks (\cite{peng2016fine}). The former relies more on high-frequency features \emph{e.g.} texture, to be successful, as opposed to low-frequency features, including the colour of objects (\cite{wang2022fine}). Additionally, high-frequency features tend to vanish in blurry or noisy images, leading to the model needing adequate information for making a correct prediction (\cite{hsu2022pedestrian}).

Image Quality Assessment (IQA) methods are commonly adopted to produce a score for an image that approximates human perception (\cite{ding2020image}). These methods can be divided into three categories: full-reference (\cite{sara2019image}), where a score is generated given two images: the original image and its distorted version; reduced-reference (\cite{han2016reduced}), where a score is given based on partially unseen information between the image pair, and no-reference (NR) (\cite{BRISQUE}), where a score is delivered for a lone image which has not been distorted. In the context of image classification, NRIQA tends to help explain the predictions given by the classification model (\cite{XU202177}).

% In general, image quality can be enhanced through super-resolution (\cite{gankhuyag2023lightweight}), deblurring and denoising (\cite{zuo2022idea}), dehazing and deraining (\cite{yang2020advancing}) etc. Those methods help to remove or suppress the low-quality features that can cause CNNs to struggle when classifying images (\cite{Dodge:2016}). This, however, only sometimes works as high-frequency features may be altered by the heavy augmentation process, which is vital for the model's efficiency.

On one hand, scores of low-quality images can be enhanced through super-resolution (\cite{gankhuyag2023lightweight}), deblurring and denoising (\cite{zuo2022idea}), dehazing and deraining (\cite{yang2020advancing}) etc. These methods help to remove or suppress the low-quality features that can cause CNNs to struggle when classifying images (\cite{Dodge:2016}). However, this is heavily dependent on the availability of high-frequency features.
% This, however, only sometimes works as high-frequency features may be altered by the heavy augmentation process, which is vital for the model's efficiency.

On the contrary, NRIQA can help alleviate the problems caused by low-quality images by being used either to screen images before being classified to guarantee the required quality has been satisfied or at the point of training to ensure the dataset includes a range of different quality images to create a robust model. The former is only applicable in the situation where the image can be retaken, but the latter can be applied in most if not all, fine-grained image classification tasks (\cite{app12010101}).

Many NRIQA methods are usually flawed by correlations with augmentation configurations independent of image quality. These configurations include image rotation, blurring, and cropping. Among many fine-grained image classification tasks, it is a high degree of difficulty to standardise images by these configurations. Thus, a thorough selection process is needed to select one or more NRIQA methods that are independent of image rotation and image size.

The NRIQA methods have been largely investigated with image focus (\cite{PERTUZ20131415}) and distortions (\cite{MRIIQA}). Additionally, low-quality images could deteriorate CNN testing, where blurred images have the most negative effect on classification models among several image deenhancement methods (\cite{Dodge:2016}). \citeauthor{wang2020deep} propose a flexible module that can be plugged into different models to cope with low-quality images and alleviate the difficulty when the IQA justification criteria are not defined. Meanwhile, \citeauthor{percep} survey current state-of-the-art IQA methods, where the aforementioned image augmentation configurations are not explored.

This work aims to discriminate between the outer packaging of commercial products made by two manufacturers in various environments with tiny differences; even domain experts struggle to distinguish them in the region of interest of the outer packaging. The challenging factors include varying lighting conditions, camera motions, mobile phone types, etc, resulting in a wide spectrum of image qualities. Concretely, our contributions are summarised as follows:

% The task in which this work will aid is to discriminate between the outer packaging of commercial products made by two manufacturers based on mobile phone images of printed codes on the underside of the bottles. The printed codes have tiny differences caused by using different industrial printers. The differences are so minute that even experts struggle to distinguish the differences between the printings. This reveals that minimal augmentations can be applied to the images. Images are taken by different models of mobile phones and in various scenarios, ranging from the production line to laboratories. These factors lead to a wide range of image qualities within the dataset. Several NRIQA methods will be adopted in screening images in an uncontrolled environment and training a robust model. Concretely, our contributions are summarised as follows:

\textbf{1)} We select an optimal set of NRIQA methods based on their robustness to external image augmentations and ability to score low-quality images correctly.

\textbf{2)} We propose a two-step mechanism to select the most discriminant subset of a given dataset to improve fine-grained image classification performance by connecting NRIQA methods and model confidence.

\textbf{3} We show that a selected training set of high-quality images leads to competitive classification performance with an acceptable degree of space and time complexity.

\textbf{4)} We adopt a varying percentage of low-quality images to demonstrate how the training set's quality should reflect the testing set's quality.

The rest of the paper is organised as follows. Section \ref{sec:materials} details the materials required to construct our framework in Section \ref{sec:methods}. Experimental results and discussions are presented in Section \ref{sec:res}. Section \ref{sec:con} summarises the work with future directions pointed out.

\section{Materials} \label{sec:materials}

\subsection{Datasets}

\begin{figure}[h]
  \centering
  \includegraphics[width=0.98\columnwidth]{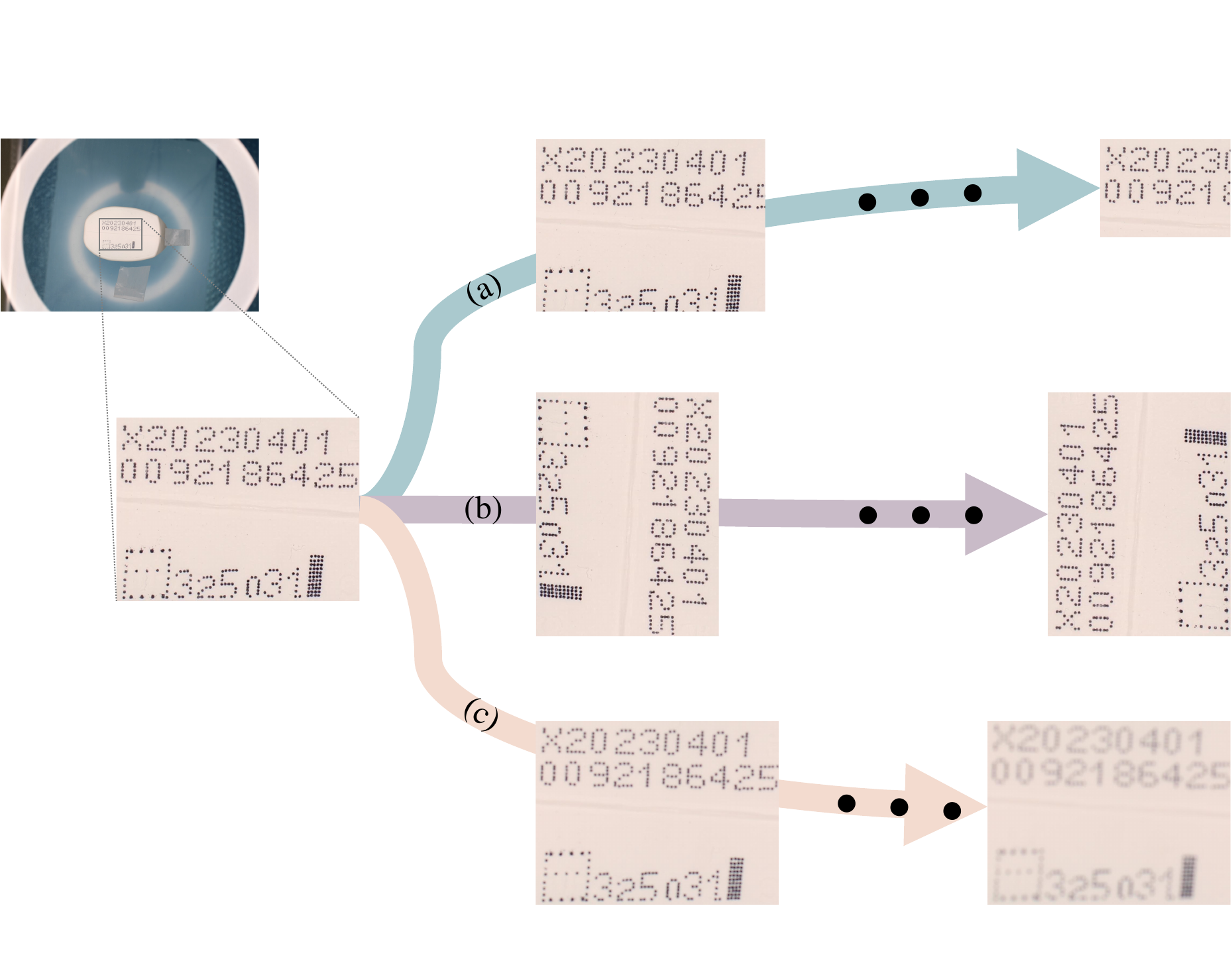}
  \caption{A raw sample of the underside of a commercial product in Dataset 1, with printed codes shown on the bottom of the bottle, is to be progressively processed through three typical image augmentation configurations: (a) cropping, (b) rotating and (c) blurring.}
   \label{fig:dataset}
\end{figure}

A total of three datasets are adopted in this work. All the datasets contain similar patterns of images but with no intersections. This means we can tune the NRIQA-guided CPS process to solve our problem without over-fitting the system to our dataset. Information about the datasets follows:

Dataset 1 contains 140 images of printed codes of the bottom of the outer packaging of commercial products, each with a resolution of $1024\times 1024$ pixels. These 140 images are of 14 distinct samples, each photographed ten times. The camera that took the photos was first correctly focused on the image and then deliberately moved out of focus by a set increment nine times by adjusting the camera's ISO setting. This created ten images of the same sample but with gradually worse quality. The images were then cropped around the printed code, similarly to how they would be in the classification model. This dataset will be used to test the effectiveness of the NRIQA methods. Figure \ref{fig:dataset} shows images of varying augmentation methods.

% shows an example of one of these images and how it is augmented to test the robustness of the NRIQA methods.

Dataset 2 includes 700 images of the bottoms of the outer packaging of commercial products, each with a resolution of $1024\times 1024$ pixels and tightly cropped around a printed dot matrix code. The class labels of the images refer to which manufacturer made the outer packaging of the commercial product. Dataset 2 is used to find the cut-off point in image quality where the classification model struggles to work effectively, as depicted in the two-step mechanism for subset selection within Figure \ref{fig:pipeline_zheming}.

Dataset 3 contains 2800 RGB images of the bottom of the outer packaging of commercial products with the same settings as Dataset 2. Dataset 3 is temporally consecutive to Dataset 2. These two datasets are independent and do not contain any identical images. Dataset 3 is used to train new models to evaluate the two-step discrimination mechanism.

% \begin{figure}[h]
%   \centering
%   \includegraphics[width=0.9\columnwidth]{images/acf.JPG}
%   \caption{Sample tightly cropped image of a printed dot matrix on the underside of a shampoo bottle.}
%    \label{fig:acf}
% \end{figure}

% \begin{figure}[h]
%   \centering
%   \includegraphics[width=0.8\linewidth]{images/blurred.png}
%   \caption{Example of sample image progressively defocused. Top-left: most in-focus image, Bottom-right: least in-focus image.}
%    \label{fig:blurred}
% \end{figure}

One of the clear images from Dataset 1 has been rotated by 15\degree\ increments to create six images ranging from the cross being rotated 0\degree\ to 75\degree. This has been done to test the robustness of the NRIQA methods to rotation in images.

One of the clear images from Dataset 1 has been cropped from the bottom and the right incrementally to create images of different resolutions, which should be of the same quality. The images were cropped by 1/20th of the original images' height and width each time, nine times over, to create ten different images. The smallest image is 1/4 of the resolution of the original image. This was done to test the robustness of each NRIQA method to image resolution.

\subsection{Image Quality Assessment Methods} \label{sec:iqa_methods}

\textbf{Statistical based NRIQA methods:}

\textbf{Variance of the Laplacian (LAPV).} The Laplace filter is commonly employed for edge detection in images. The quality of an image can be measured by taking the Laplacian variance of the image (\cite{PechLAPLACE:2000}).

\textbf{Modified Laplacian (LAPM).} \citeauthor{NayerMODLAPLACE:1994} suggested a focus measure based on an alternative definition of Laplacian, which can be used as an NRIQA.

\textbf{Sum of the Wavelet Coefficients (WAVS).} Discrete Wavelet Transform (DWT) of an image could be helpful where, in its first level, the image is decomposed into four sub-images. \citeauthor{YangEtAl:2003} calculated the corresponding horizontal, vertical, and diagonal coefficients.

\textbf{BRISQUE.} \citeauthor{BRISQUE} propose a Blind/Referenceless Image Spatial Quality Evaluator (BRISQUE) using natural scene statistics (NSS) in the spatial domain (\cite{BRISQUE}). NSS is designed to quantify losses of ``naturalness'' in images.

% does not use distortion-specific features, \emph{e.g.} blurring or ringing, and instead uses the NSS to quantify possible losses of ``naturalness'' in images.

\textbf{NIQE.} \citeauthor{NIQE} developed on BRISQUE (\cite{BRISQUE}) to create a Natural Image Quality Evaluator (NIQE) (\cite{NIQE}). NIQE used the same NSS as BRISQUE, but instead of basing results on the features of distorted images or images perceived as low-quality by human perception, NIQE uses measurable deviations from statistical regularities observed in natural images.

\noindent{\textbf{Deep Learning Based NRIQA Methods:}}

\textbf{MUSIQ.} \citeauthor{MUSIQ} propose a MUltiScale Image Quality transformer (MUSIQ) to calculate a quality score for an image (\cite{MUSIQ}). MUSIQ is pretrained on ImageNet and then fine-tuned for use on large-scale image quality datasets (\cite{ying2019patches}, \cite{9156490}, \cite{Hosu_2020}). MUSIQ does not require a fixed shape; thus, less image augmentation may lead to a more significant change in image quality.
% MUSIQ does not have fixed shape constraints; thus, images do not have to be resized to fit the model. This means there is less image augmentation in use, which may lead to a more significant change in image quality.

\textbf{MANIQA.} \citeauthor{maniqa} presented a Multidimensional Attention Network for No-Reference Image Quality assessment (MANIQA) (\cite{maniqa}). MANIQA is different from MUSIQ as it works on GAN-based distortion images merely. MANIQA is pretrained on the PIPAL dataset (\cite{gu2020pipal}) that contains a mixture of natural and GAN-based images.

\textbf{HyperIQA.} \citeauthor{hyperiqa} adopted hypernetworks based solely on images with naturally occurring distortion (\cite{hyperiqa}). Experimental results indicate that it performs well for natural and synthetic distortions. The model is pretrained on KonIQ-10k (\cite{Hosu_2020}).

\subsection{Deep Neural Classifiers}

We employ a ResNet34 model (\cite{he2015deep}) trained on a dataset of 3090 RGB images of the bottom of the outer packaging of commercial products (\cite{jackson2020camera}) as the baseline model for this work. We adopt the Adam optimiser for model training, a learning rate of $10^{-4}$, a weight decay of 0, and the categorical cross-entropy loss function. This model is used for finding the cut-off point for acceptable image quality by looking at the model's confidence for the images in Dataset 2.

We also utilise pretrained models of AlexNet (\cite{krizhevsky2012imagenet}), ResNet18, ResNet34 (\cite{he2015deep}), and VGG19 (\cite{Simonyan15}) to evaluate the effect image quality has on training. The models are pretrained on ImageNet (\cite{russakovsky2015imagenet}).

% \begin{figure}[h]
%   \centering
%   \includegraphics[width=0.8\linewidth]{images/resize.png}
%   \caption{Example of sample image progressively cropped. Top-left: original sample image, Bottom-right: Most cropped image.}
%    \label{fig:resized}
% \end{figure}

\section{Methods} \label{sec:methods}

\begin{figure*}[!ht]
  \centering
  \includegraphics[width=0.98\linewidth]{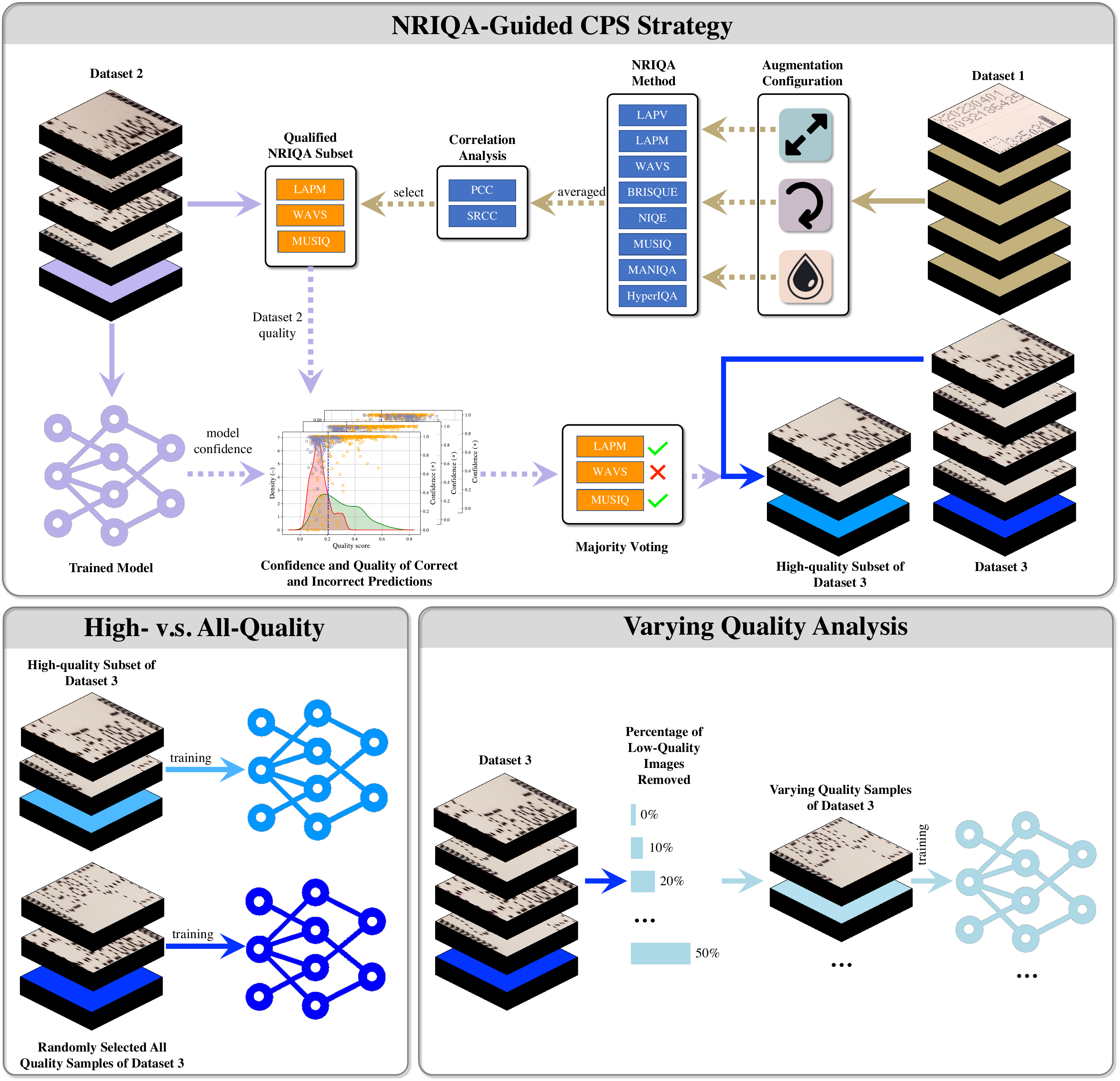}
  \caption{Overall framework of the proposed two-step mechanism of high-quality image subset selection of the image dataset for improved fine-grained image classification. The upper row depicts the process of seeking the most appropriate subset of NRIQA methods specified in Section \ref{sec:iqa_methods} and the majority voting procedure of selecting high-quality images from a given dataset specified in Section \ref{sec:cpsmeth}. The bottom left box denotes how we used training sets made up of only high-quality and mixed-quality images to show how the accuracy of a model is affected by the quality of its training set in Section \ref{sec:exp1meth}. The bottom right box denotes how we employed a varying amount of low-quality images in the training set to find the optimal mix of low and high-quality images in Section \ref{sec:exp2meth}.}
   \label{fig:pipeline_zheming}
\end{figure*}

In this section, we detail the framework of the NRIQA-guided CPS process of image quality within a dataset of images depicted in Figure \ref{fig:pipeline_zheming}. Specifically, the correlation metrics used to evaluate the effectiveness and robustness of the NRIQA methods are detailed in Section \ref{sec:cor}. This is depicted in the top row of the top box of Figure \ref{fig:pipeline_zheming}. Section \ref{sec:cpsmeth} details how we decide the values for the cut-off points of the NR-

\noindent IQA methods for image quality as pictured on the bottom row of the top box in Figure \ref{fig:pipeline_zheming}. Section \ref{sec:exp1meth} details how we used training sets made up of only high-quality images and mixed-quality images to show how the accuracy of a model is affected by the quality of its training set. This is drawn in the bottom left box of Figure \ref{fig:pipeline_zheming}. Our further experiments with varying quality are detailed in Section \ref{sec:exp2meth}, revealing how we employed various low-quality images in the training set to find the optimal mix of low and high-quality images. This is visualised in the bottom right box of Figure \ref{fig:pipeline_zheming}.

\subsection{NRIQA-Guided CPS Strategy}

\subsubsection{Correlation Metrics of NRIQA Methods}
\label{sec:cor}

Given the qualities of images measured by several assessment metrics detailed in Section \ref{sec:iqa_methods}, we then, as depicted in the upper of Figure \ref{fig:pipeline_zheming}, determine the qualified NRIQA metrics through correlation analysis. 

\textbf{Pearson Correlation Coefficient (PCC)} measures the degree of relationship between a pair of variables (\cite{adler2010quantifying}). It helps to explain the correlations in the results of our experiments. Specifically, PCC is calculated by:

\begin{equation}
    \textnormal{PCC} (I_{\textnormal{idx}}, I_{\textnormal{qvc}}) = \frac{COV (I_{\textnormal{idx}}, I_{\textnormal{qvc}})}{SD(I_{\textnormal{idx}}) * SD(I_{\textnormal{qvc}})},
\end{equation}
where $I_{\textnormal{idx}}$ denotes the index of the image and $I_{\textnormal{qvc}}$ represents the image quality score, $COV(\cdot, \cdot)$ and $SD(\cdot)$ correspond to covariance and standard deviation, respectively.

\textbf{Spearman Rank Correlation Coefficient (SRCC)}  reveals the correlation of the order of results rather than the value of the results (\cite{lyerly1952average}). SRCC is calculated as follows:

\begin{equation}
    \textnormal{SRCC} (I_{\textnormal{idx}}, I_{\textnormal{qvc}}) = \textnormal{PCC}(\textnormal{rank}(I_{\textnormal{idx}}),\textnormal{rank}(I_{\textnormal{qvc}})),
\end{equation}
where $rank(\cdot)$ denotes the operation of ordering the results from smallest to largest.

It is a common choice to adopt both PCC and SRCC in practice (\cite{cheng2021brain}), as the former gives a general idea of the gradient of the relationship between the image index in the dataset and its corresponding quality score. In contrast, the latter indicates which methods correlate strongly with the order of images, \emph{e.g.} the clearest image to the blurriest image.

To evaluate the robustness and efficiency of the NRIQA set, images from each of the augmentations of Dataset 1 are run through the eight quality metrics. The SRCC and the PCC are calculated from the averaged quality scores to show the correlation between image augmentation configurations and the NRIQA scores produced. Experimental setups are detailed below.

% The dataset of progressively cropped images is used to justify how robust each NRIQA method is to image rotation. A survived NRIQA method should have a low correlation between the NRIQA scores returned and the image's resolution. Similarly, the dataset of images of rotated images is used to verify the robustness of each NRIQA method to image rotation. A survived NRIQA method, which is less susceptible to rotation, will have a low correlation between the NRIQA score and the rotation of the image. Lastly, the dataset of clear to blurry images, specified in the previous section, is used to test how effective each NRIQA method is at scoring blurry images. A high correlation between the blurriness of the image and the NRIQA score is expected from a survived NRIQA method, as the scores should be directly correlated with the quality of the images. The correlation coefficients will be averaged across the 14 samples.
The progressively cropped images from Dataset 1 justify the robustness of each NRIQA method to image rotation. A surviving NRIQA method exhibits a low correlation between returned scores and image resolution. Similarly, the rotated images assess each NRIQA method's robustness to rotation, whereas surviving methods display a low correlation between scores and image rotation. Lastly, the clear-to-blurry images from Dataset 1 gauge the effectiveness of each NRIQA method in scoring blurry images. A surviving NRIQA method should exhibit a high correlation between image blurriness and scores, reflecting score-quality alignment. Correlation coefficients will be averaged across 14 samples.

The above results will inform which NRIQA methods should be selected to restrict low-quality images from being used in the classification system. When multiple methods are chosen, the common majority voting methods will be triggered, \emph{e.g.} if the majority of methods produce low scores, the image will be deemed ``blurry'' and will not be used.

\subsubsection{Cut-off Point Determination for Subset Selection}
\label{sec:cpsmeth}

To validate the correlation between image quality and the accuracy of a fine-grained classification model, the confidence of a pretrained ResNet34 is compared to the image quality scores generated by the NRIQA methods. A high, positive correlation between confidence and image quality suggests that the model does struggle with lower-quality images, as suspected. To this end, we run the entirety of Dataset 2 through the current model and check the confidence and image quality values produced.

The cut-off points for each NRIQA method are determined by looking at the image quality distribution over correct and incorrect predictions. By calculating the Kernel Density Estimation (KDE) (\cite{chen2002robust}) of the distributions, we can observe where the two curves cross and use this as a sensible value to restrict lower-quality images. 

\begin{table*}[!ht]
    \centering
    \resizebox{1\textwidth}{!}{
    \begin{tabular}{llcccccccc}
        \toprule
        \multirowcell{2}{Augumentation\\Configuration} & \multirowcell{2}{Correlation\\ Metric} & \multicolumn{8}{c}{NRIQA Method}\\
        \cmidrule{3-10}
        && LAPV  & LAPM           & WAVS  & BRISQUE & NIQE  & MUSIQ           & MANIQA  & HyperIQA  \\ \midrule
        \multirow{2}{*}{cropping} & PCC$\downarrow$ & 0.337 & 0.322          & 0.345 & 0.599 & 0.774 & 0.954          & 0.937 & \textbf{0.150} \\
        & SRCC$\downarrow$ & 0.358 & \textbf{0.152} & 0.164 & 0.406 & 0.969 & 0.952          & 0.927 & 0.333 \\
        \midrule
        \multirow{2}{*}{rotating} &PCC$\downarrow$  & 0.406 & 0.570 & \textbf{0.050} & 0.336 & 0.279 & 0.316 & 0.062 & 0.133 \\
        &SRCC$\downarrow$ & 0.486 & 0.486 & \textbf{0.086} & 0.200 & 0.200 & 0.143 & 0.200 & 0.314 \\
        \midrule
        \multirow{2}{*}{blurring} & PCC$\uparrow$  & 0.770          & 0.878          & 0.845 & 0.855 & 0.053          & \textbf{0.907} & 0.695 & 0.169 \\
        & SRCC$\uparrow$ & \textbf{1.000} & 0.996          & 0.973 & 0.965 & 0.165          & 0.885          & 0.626 & 0.194 \\         
        \bottomrule
    \end{tabular}}
    \caption{Average correlation coefficients for image augmentation configurations. Bold values show the most desirable result in each row. $\downarrow$ denotes where a lower correlation is favourable, whereas the $\uparrow$ indicates where a higher correlation is preferable.}
    \label{tab:corr_scores_overall_zheming}
\end{table*}

\noindent This value should represent the point at which the image is too blurry for the model to classify the image reliably.

\subsection{High- v.s. All-Quality}
\label{sec:exp1meth}

To assess the model's performance on high-quality images, we utilised the specified NRIQA thresholds to form a subset of Dataset 3. This subset exclusively contained high-quality images. 10\% were designated for testing, 10\% for evaluation, and the rest for training a manufacturer classification model based on outer packaging of commercial products underside images. Four model architectures were used (AlexNet, ResNet18, ResNet34, and VGG19). The mean and standard deviation of five runs are evaluated.

To compare the aforementioned model with one trained on diverse image quality, another model was trained for the same task but using images across all the quality levels. Random images were selected from Dataset 3 to match the high-quality image subset's size. This approach ensured fairness by avoiding any subset advantage through data volume discrepancy.

All trained models undergo testing using the high-quality test set. This comparison will reveal the superior model for high-quality image classification. As future image acquisition will be limited to high-quality images, this approach accurately assesses each model's effectiveness.

\subsection{Varying Quality Analysis}
\label{sec:exp2meth}

A varying range of image qualities was incorporated to review how the model's accuracy changes as the quality of the images increases or decreases. A random subset of Dataset 3 was selected. Different percentages ranging from 0\% to 50\% of the lowest quality images in the subset were removed to create multiple subsets of various ranges of quality. The number of images in each subset was standardised by selecting a larger original subset depending on the number of images removed. A ResNet34 model was trained five times on each subset and tested on the high-quality testing set from the experiment above. The model was trained using the same configuration as in Section \ref{sec:exp1meth}. The averages and standard deviations for the five runs were evaluated in Section \ref{sec:varres}.

\section{Results} \label{sec:res}

\subsection{NRIQA Method Assessments}

Table \ref{tab:corr_scores_overall_zheming} summarises the correlation coefficients of PCC and SRCC over the averaged scores of each NRIQA method, where the preferable PCC and SRCC are marked in bold. 

\subsubsection{Cropped Images}

From the results reported, the LAPM and HyperIQA had the lowest correlation with the image resolutions and, thus, are the most robust methods to image cropping. In contrast, NIQE, MANIQA, and MUSIQ are relatively less robust models for image cropping.

\subsubsection{Rotated Images}

As for the results reported for image rotation in Table \ref{tab:corr_scores_overall_zheming}, WAVS had the lowest correlation to rotation and is more robust compared against the rest of the NRIQA methods to image rotation.

\subsubsection{Blurred Images}

Regarding image blurring, LAPV and MUSIQ had the highest correlation between the NRIQA score and the intentional defocus of the image. Deep learning-based methods, other than MUSIQ, had lower correlations. This is due to the images of the outer packaging of commercial products being markedly distinct from the images the models were pretrained on.

\begin{figure*}[!ht]
  \centering
  \includegraphics[width=\linewidth]{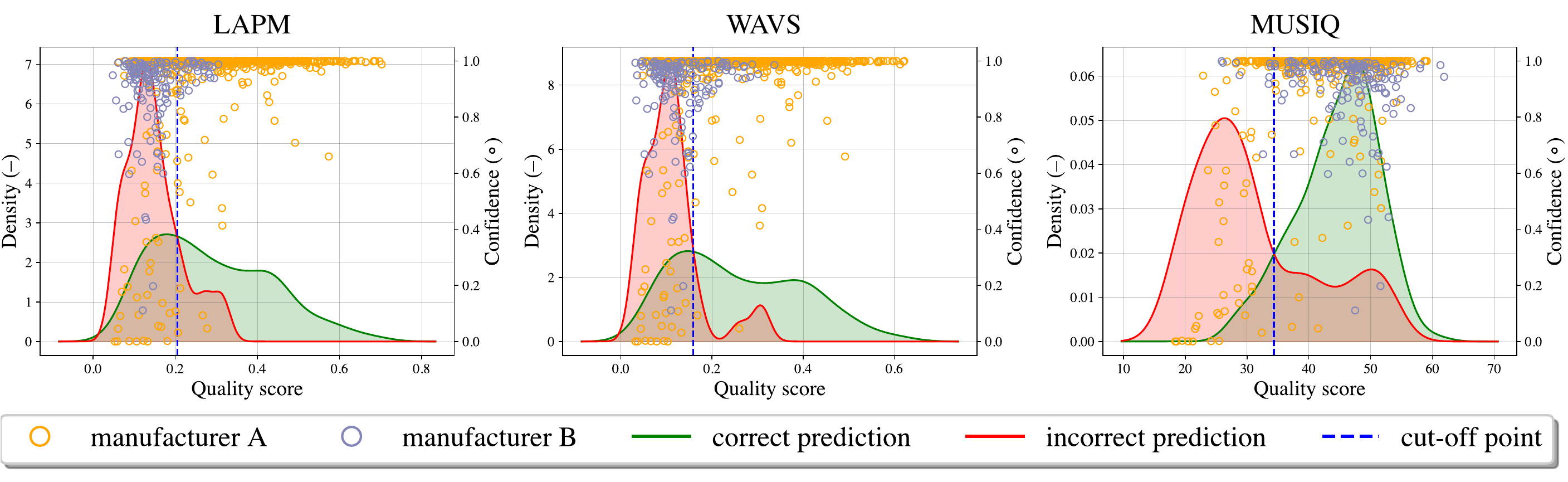}
  \caption{The scatter plots reveal the relationship between the image-wise quality score in the $x$-axis and the prediction confidence of the pretrained classification model in the $y$-axis for images in Dataset 2, as well as the corresponding density distribution of correct/incorrect predictions.}
   \label{fig:conf_density_zheming}
\end{figure*}

\begin{figure*}[!ht]
  \centering
  \includegraphics[width=\linewidth]{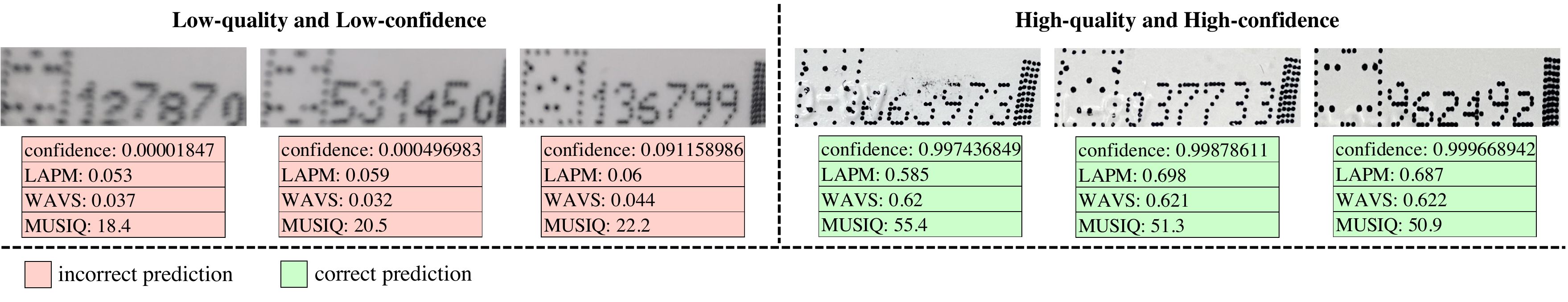}
  \caption{Explanations of the model predictions on low- and high-quality testing images.}
   \label{fig:hi_versus_low_zheming}
\end{figure*}

From there, we determined to use a combination of the LAPM, WAVS, and MUSIQ in a voting system for the remaining experiments.

\subsection{Cut-off Point Determination for Subset Selection}
After running the whole of Dataset 3 through the classification model, the model achieved 95\% accuracy. Figure \ref{fig:conf_density_zheming} depicts the confidence of the classification model against the quality scores generated for the image for each of the NRIQA methods. All three graphs show a clear, strong correlation between the confidence of the classification model and the quality scores generated by the NRIQA methods. Most misclassifications happen within the lower half of image quality for all three NRIQA methods. This shows that image quality strongly affects the classification model's ability to classify images accurately.

% \begin{figure}[htbp!]
%   \centering
%   \begin{subfigure}{0.5\textwidth}
%     \includegraphics[width=\linewidth]{images/scatter_wavs.pdf}
%     \caption{Sum of the Wavelet Coeffcients}\label{fig:scattera}
%   \end{subfigure}
%   \begin{subfigure}{0.5\textwidth}
%     \includegraphics[width=\linewidth]{images/scatter_lapm.pdf}
%     \caption{Modified Laplacian} \label{fig:scatterb}
%   \end{subfigure}
%   \begin{subfigure}{0.5\textwidth}
%     \includegraphics[width=\linewidth]{images/scatter_musq.pdf}
%     \caption{MUSIQ} \label{fig:scatterc}
%   \end{subfigure}
%   \caption{Scatter graphs showing relationship between image quality and the confidence of the classification model. Any confidence below 0.5 is a misclassification. Blue markers denote images of class Manufacturer A. Orange markers denote images of class Manufacturer B.}
%    \label{fig:scatters}
% \end{figure}

Cut-off points for each of the three surviving methods were suggested by looking at the quality of images where the classification confidence drops and the model starts to misclassify images. Figure \ref{fig:conf_density_zheming} reveals the KDEs of image quality for correct and incorrect classifications for each of the selected NRIQA methods. Specifically, the cut-off points set were 0.206 for LAPM, 0.159 for WAVS, and 34.4 for MUSIQ. These values are used in the voting system to determine whether an image is to be selected. Then, the voting system will reject the image if it produces low-quality scores for at least two of the three surviving NRIQA methods; in line with Figure \ref{fig:hi_versus_low_zheming}.

% \begin{figure}[htbp]
%   \centering
%   \begin{subfigure}{0.5\textwidth}
%     \includegraphics[width=\linewidth]{images/histo_wavs.pdf}
%     \caption{Sum of the Wavelet Coefficients}\label{fig:histoa}
%   \end{subfigure}
%   \begin{subfigure}{0.5\textwidth}
%     \includegraphics[width=\linewidth]{images/histo_lapm.pdf}
%     \caption{Modified Laplacian} \label{fig:histob}
%   \end{subfigure}
%   \begin{subfigure}{0.5\textwidth}
%     \includegraphics[width=\linewidth]{images/histo_musq.pdf}
%     \caption{MUSIQ} \label{fig:histoc}
%   \end{subfigure}
%   \caption{KDEs showing distribution of image quality for correctly and incorrectly classified images. The green lines show the crossing points of the distributions and are the cut-off points used for the three NRIQA methods.}
%    \label{fig:histos}
% \end{figure}

\subsection{High v.s. All-Quality}

% Old version
%Table \ref{tab:tests} reports the average testing accuracies of the models trained on high-quality and all-quality images. All models were tested on a data set made up of only high-quality data from Dataset 3 unseen to the models, as these will be the most similar images to what the system will be employed for in the real world once the quality thresholds are implemented. The models trained on only high-quality images achieved an average final accuracy of \textbf{85.4\%}. The models trained on the full range of quality images achieved an average final accuracy of \textbf{81.2\%}. These final accuracies show that when the model is trained only on high-quality images, it performs on average \textbf{4.2\%} better when tested on high-quality images. This is a small increase, but when factored into a large system, this is an incremental step towards better accuracy.

Table \ref{tab:tests} reports the average testing accuracies of the models trained on high-quality and all-quality images. All models were tested on a data set of only high-quality data from Dataset 3 unseen to the models, as these will be the most similar images to what the system will be employed on in an uncontrolled environment once the quality thresholds are implemented. 

\begin{table}[!ht!]
\centering
\resizebox{\columnwidth}{!}{%
    \begin{tabular}{llcc}
        \toprule
        Model        & Quality         & Accuracy (\%) &         AUC (\%)\\
        \midrule

        \multirow{2}{*}{AlexNet} & High-quality                 
                      & \textbf{79.3$\pm$3.3}  & 96.1$\pm$1.6\\
        
                 & All-quality      
                      & 78.4$\pm$4.2  & 96.6$\pm$1.9\\

        \midrule

        \multirow{2}{*}{VGG19}        & High-quality                 
                      & \textbf{83.3$\pm$4.2}  & 96.3$\pm$1.5\\
        
                     & All-quality  
                      & 82.6$\pm$6.7  & 95.2$\pm$2.0\\

        \midrule
        
        \multirow{2}{*}{ResNet18}     & High-quality                 
                      & \textbf{86.6$\pm$2.0}  & 94.4$\pm$1.0\\
        
                     & All-quality  
                      & 85.1$\pm$2.9  & 94.7$\pm$0.5\\

        \midrule
        
        \multirow{2}{*}{ResNet34}     & High-quality 
                      & \textbf{85.4$\pm$3.1}  &
        94.8$\pm$1.2\\
        
                     & All-quality  
                      & 81.2$\pm$2.1  &  
        93.2$\pm$1.3\\

        \bottomrule
    \end{tabular}}
    \caption{Testing results for models trained on high- and all-quality images in five runs with optimal accuracy marked in bold. Each model is trained on 1293 images with 40 epochs.}
    \label{tab:tests}
\end{table}

In all cases, the models trained on only high-quality images outperformed the models trained on the full range of image quality. The architecture with the highest difference in accuracy between high and all-quality images was ResNet34, which had an average accuracy of 85.4\% when trained on only high-quality images, compared to 81.2\% when trained on the full range of image quality. This is only a small increase of 4.2\%, but when factored into a large system, this is an incremental step towards better accuracy.

\subsection{Varying Quality Analysis}
\label{sec:varres}

Table \ref{tab:percents} reports the results for the experiment described in Section \ref{sec:exp2meth}. The results indicate that the accuracy of the model increases as the percentage of removed images approaches 30\% and then drops slightly and levels off afterwards. 

\begin{table}[!ht]
\centering
    \begin{tabular}{ccc}
        \toprule
        Percent Removed (\%)   & Accuracy (\%) & AUC (\%)     \\
        \midrule
        0                                               & 81.5$\pm$4.4  & 96.9$\pm$1.7 \\
        10                                               & 82.0$\pm$2.6  & 96.7$\pm$1.1 \\
        20                                            & 81.9$\pm$5.1  & 95.4$\pm$0.8 \\
        30                                               & \textbf{84.1$\pm$3.5}  & 96.8$\pm$1.5 \\
        40                                                & 82.4$\pm$3.2  & 95.1$\pm$2.6 \\
        50                                               & 82.6$\pm$4.5  & 94.5$\pm$2.4 \\
        \bottomrule
    \end{tabular}
    \caption{Averaged results of ResNet34 trained on various combinations of high- and low-quality images for five runs with the optimal accuracy marked in bold. Each model is trained on 1293 images with 40 epochs.}
    \label{tab:percents}
\end{table}

This might be caused by the quality of the training set at 30\% removed being the most representative of the testing set. This shows the importance of having a training set that is reflective of the data from an uncontrolled environment and that using images in training of lower quality than expected in the uncontrolled environment does not improve the accuracy of the model.

\subsection{Discussions}

From our investigation, WAVS, LAPM, and MUSIQ are the most effective NRIQA methods with the smallest degree of correlation to image rotation or cropping. We observe that using a combination of these methods in a voting system allows them to compensate for each other's weaknesses, \emph{i.e.} MUSIQ correlating image resolution while otherwise being a very effective NRIQA method.

We evaluated our chosen NRIQA methods and looked at the correlation between image quality and fine-grained classification accuracy by examining the existing model's confidence in images of different quality scores. We found a strong correlation between image quality scores and the models' confidence in correct classifications across all three selected NRIQA methods. This backs up our hypothesis that low image quality does reduce a fine-grained image classification model's ability to classify images accurately.

We then looked at how image quality affects model training. The results show that a model trained exclusively on high-quality performs better at classifying high-quality images than a model trained on the full range of image quality. This is useful as it helps direct how we augment data for a task with a guaranteed range of quality. The results suggest that augmenting images using blurring past the guaranteed range of image quality may not be useful as a pre-processing step in training.

We found that a mixture of low- and high-quality images can be used as long as 30\% of the lowest-quality images are removed. This is useful as it helps validate the CPS strategy. A similar process could also be used in other problems to help find the optimal subset of images for training within a given dataset.

\section{Conclusions} \label{sec:con}

In this paper, we propose a two-step mechanism to inform the process of selecting the high-quality subset of a given dataset by taking both the density distribution of averaged quality scores of qualified NRIQA methods and the prediction confidence of a deep neural network into consideration. The models trained on the high-quality images outperform the ones trained on both high- and low-quality images. Additionally, the high-quality selected using the proposed framework has been proven to be compatible with partial low-quality ones with acceptable classification accuracy achieved. Though competitive results have been obtained, investigations of the proposed two-step mechanism on larger-scale datasets could be treated as active future work.

\bibliographystyle{apalike}
{\small
\bibliography{example}}

\begin{thebibliography}{}

\bibitem[Adler and Parmryd, 2010]{adler2010quantifying}
Adler, J. and Parmryd, I. (2010).
\newblock Quantifying colocalization by correlation: the pearson correlation coefficient is superior to the mander's overlap coefficient.
\newblock {\em Cytometry Part A}, 77(8):733--742.

\bibitem[Aqqa et~al., 2019]{aqqa2019understanding}
Aqqa, M., Mantini, P., and Shah, S.~K. (2019).
\newblock Understanding how video quality affects object detection algorithms.
\newblock In {\em International Conference on Computer Vision Theory and Applications}, pages 96--104.

\bibitem[Chen and Meer, 2002]{chen2002robust}
Chen, H. and Meer, P. (2002).
\newblock Robust computer vision through kernel density estimation.
\newblock In {\em European Conference on Computer Vision}, pages 236--250.

\bibitem[Cheng et~al., 2021]{cheng2021brain}
Cheng, J., Liu, Z., Guan, H., Wu, Z., Zhu, H., Jiang, J., Wen, W., Tao, D., and Liu, T. (2021).
\newblock Brain age estimation from {MRI} using cascade networks with ranking loss.
\newblock {\em IEEE Transactions on Medical Imaging}, 40(12):3400--3412.

\bibitem[Deng et~al., 2009]{russakovsky2015imagenet}
Deng, J., Dong, W., Socher, R., Li, L.-J., Li, K., and Fei-Fei, L. (2009).
\newblock Imagenet: A large-scale hierarchical image database.
\newblock In {\em IEEE Conference on Computer Vision and Pattern Recognition}, pages 248--255.

\bibitem[Ding et~al., 2020]{ding2020image}
Ding, K., Ma, K., Wang, S., and Simoncelli, E.~P. (2020).
\newblock Image quality assessment: Unifying structure and texture similarity.
\newblock {\em IEEE Transactions on Pattern Analysis and Machine Intelligence}, 44(5):2567--2581.

\bibitem[Dodge and Karam, 2016]{Dodge:2016}
Dodge, S.~F. and Karam, L. (2016).
\newblock Understanding how image quality affects deep neural networks.
\newblock In {\em International Conference on Quality of Multimedia Experience}, pages 1--6.

\bibitem[Fang et~al., 2020]{9156490}
Fang, Y., Zhu, H., Zeng, Y., Ma, K., and Wang, Z. (2020).
\newblock Perceptual quality assessment of smartphone photography.
\newblock In {\em IEEE Conference on Computer Vision and Pattern Recognition}, pages 3674--3683.

\bibitem[Gankhuyag et~al., 2023]{gankhuyag2023lightweight}
Gankhuyag, G., Yoon, K., Park, J., Son, H.~S., and Min, K. (2023).
\newblock Lightweight real-time image super-resolution network for 4k images.
\newblock In {\em IEEE Conference on Computer Vision and Pattern Recognition}, pages 1746--1755.

\bibitem[Gu et~al., 2020]{gu2020pipal}
Gu, J., Cai, H., Chen, H., Ye, X., Ren, J., and Dong, C. (2020).
\newblock Pipal: a large-scale image quality assessment dataset for perceptual image restoration.
\newblock In {\em European Conference on Computer Vision}, pages 633--651.

\bibitem[Gupta et~al., 2021]{gupta2021deep}
Gupta, A., Anpalagan, A., Guan, L., and Khwaja, A.~S. (2021).
\newblock Deep learning for object detection and scene perception in self-driving cars: Survey, challenges, and open issues.
\newblock {\em Array}, 10:100057.

\bibitem[Ha et~al., 2022]{ha2022fine}
Ha, Y., Du, Z., and Tian, J. (2022).
\newblock Fine-grained interactive attention learning for semi-supervised white blood cell classification.
\newblock {\em Biomedical Signal Processing and Control}, 75:103611.

\bibitem[Han et~al., 2016]{han2016reduced}
Han, Z., Zhai, G., Liu, Y., Gu, K., and Zhang, X. (2016).
\newblock A reduced-reference quality assessment scheme for blurred images.
\newblock In {\em IEEE Visual Communications and Image Processing}, pages 1--4.

\bibitem[He et~al., 2016]{he2015deep}
He, K., Zhang, X., Ren, S., and Sun, J. (2016).
\newblock Deep residual learning for image recognition.
\newblock In {\em IEEE Conference on Computer Vision and Pattern Recognition}, pages 770--778.

\bibitem[Hosu et~al., 2020]{Hosu_2020}
Hosu, V., Lin, H., Sziranyi, T., and Saupe, D. (2020).
\newblock {KonIQ}-10k: An ecologically valid database for deep learning of blind image quality assessment.
\newblock {\em {IEEE} Transactions on Image Processing}, 29:4041--4056.

\bibitem[Hsu and Chen, 2022]{hsu2022pedestrian}
Hsu, W.-Y. and Chen, P.-C. (2022).
\newblock Pedestrian detection using stationary wavelet dilated residual super-resolution.
\newblock {\em IEEE Transactions on Instrumentation and Measurement}, 71:1--11.

\bibitem[Jackson et~al., 2021]{jackson2020camera}
Jackson, P.~T., Bonner, S., Jia, N., Holder, C., Stonehouse, J., and Obara, B. (2021).
\newblock Camera bias in a fine grained classification task.
\newblock In {\em IEEE International Joint Conference on Neural Networks}, pages 1--8.

\bibitem[Ke et~al., 2021]{MUSIQ}
Ke, J., Wang, Q., Wang, Y., Milanfar, P., and Yang, F. (2021).
\newblock {MUSIQ}: Multi-scale smage suality transformer.
\newblock In {\em IEEE International Conference on Computer Vision}, pages 5128--5137.

\bibitem[Krizhevsky et~al., 2012]{krizhevsky2012imagenet}
Krizhevsky, A., Sutskever, I., and Hinton, G.~E. (2012).
\newblock Imagenet classification with deep convolutional neural networks.
\newblock In {\em Advances in Neural Information Processing Systems}, pages 1106--1114.

\bibitem[Lyerly, 1952]{lyerly1952average}
Lyerly, S.~B. (1952).
\newblock The average spearman rank correlation coefficient.
\newblock {\em Psychometrika}, 17(4):421--428.

\bibitem[Mittal et~al., 2012]{BRISQUE}
Mittal, A., Moorthy, A.~K., and Bovik, A.~C. (2012).
\newblock No-reference image quality assessment in the spatial domain.
\newblock {\em IEEE Transactions on Image Processing}, 21:4695--4708.

\bibitem[Mittal et~al., 2013]{NIQE}
Mittal, A., Soundararajan, R., and Bovik, A.~C. (2013).
\newblock Making a “completely blind” image quality analyzer.
\newblock {\em IEEE Signal Processing Letters}, 20(3):209--212.

\bibitem[Nayar and Nakagawa, 1994]{NayerMODLAPLACE:1994}
Nayar, S.~K. and Nakagawa, Y. (1994).
\newblock Shape from focus.
\newblock {\em IEEE Transactions on Pattern Analysis and Machine Intelligence}, 16(8):824--831.

\bibitem[Pech-Pacheco et~al., 2000]{PechLAPLACE:2000}
Pech-Pacheco, J., Cristobal, G., Chamorro-Martinez, J., and Fernandez-Valdivia, J. (2000).
\newblock Diatom autofocusing in brightfield microscopy: a comparative study.
\newblock In {\em International Conference on Pattern Recognition}, pages 314--317.

\bibitem[Peng et~al., 2016]{peng2016fine}
Peng, X., Hoffman, J., Stella, X.~Y., and Saenko, K. (2016).
\newblock Fine-to-coarse knowledge transfer for low-res image classification.
\newblock In {\em IEEE International Conference on Image Processing}, pages 3683--3687.

\bibitem[Pertuz et~al., 2013]{PERTUZ20131415}
Pertuz, S., Puig, D., and Garcia, M.~A. (2013).
\newblock Analysis of focus measure operators for shape-from-focus.
\newblock {\em Pattern Recognition}, 46(5):1415--1432.

\bibitem[Sabbatini et~al., 2021]{sabbatini2021computer}
Sabbatini, L., Palma, L., Belli, A., Sini, F., and Pierleoni, P. (2021).
\newblock A computer vision system for staff gauge in river flood monitoring.
\newblock {\em Inventions}, 6(4):79.

\bibitem[Sara et~al., 2019]{sara2019image}
Sara, U., Akter, M., and Uddin, M.~S. (2019).
\newblock Image quality assessment through {FSIM}, {SSIM}, {MSE} and {PSNR—a} comparative study.
\newblock {\em Journal of Computer and Communications}, 7(3):8--18.

\bibitem[Sharif~Razavian et~al., 2014]{razavian2014cnn}
Sharif~Razavian, A., Azizpour, H., Sullivan, J., and Carlsson, S. (2014).
\newblock Cnn features off-the-shelf: an astounding baseline for recognition.
\newblock In {\em IEEE Conference on Computer Vision and Pattern Recognition workshops}, pages 806--813.

\bibitem[Simonyan and Zisserman, 2015]{Simonyan15}
Simonyan, K. and Zisserman, A. (2015).
\newblock Very deep convolutional networks for large-scale image recognition.
\newblock In {\em International Conference on Learning Representations}, pages 1--14.

\bibitem[Stepien and Oszust, 2022]{MRIIQA}
Stepien, I. and Oszust, M. (2022).
\newblock A brief survey on no-reference image quality assessment methods for magnetic resonance images.
\newblock {\em Journal of Imaging}, 8(6):160--178.

\bibitem[Su et~al., 2020]{hyperiqa}
Su, S., Yan, Q., Zhu, Y., Zhang, C., Ge, X., Sun, J., and Zhang, Y. (2020).
\newblock Blindly assess image quality in the wild guided by a self-adaptive hyper network.
\newblock In {\em IEEE Conference on Computer Vision and Pattern Recognition}, pages 3664--3673.

\bibitem[Varga, 2022]{app12010101}
Varga, D. (2022).
\newblock No-reference image quality assessment with convolutional neural networks and decision fusion.
\newblock {\em Applied Sciences}, 12(1):101--118.

\bibitem[Wang et~al., 2023]{wang2022fine}
Wang, M., Zhao, P., Lu, X., Min, F., and Wang, X. (2023).
\newblock Fine-grained visual categorization: A spatial--frequency feature fusion perspective.
\newblock {\em IEEE Transactions on Circuits and Systems for Video Technology}, 33(6):2798--2812.

\bibitem[Wang et~al., 2020]{wang2020deep}
Wang, Y., Cao, Y., Zha, Z.-J., Zhang, J., and Xiong, Z. (2020).
\newblock Deep degradation prior for low-quality image classification.
\newblock In {\em IEEE Conference on Computer Vision and Pattern Recognition}, pages 11049--11058.

\bibitem[Xu et~al., 2021]{XU202177}
Xu, Y., Wei, M., and Kamruzzaman, M. (2021).
\newblock Inter/intra-category discriminative features for aerial image classification: A quality-aware selection model.
\newblock {\em Future Generation Computer Systems}, 119:77--83.

\bibitem[Yang and Nelson, 2003]{YangEtAl:2003}
Yang, G. and Nelson, B.~J. (2003).
\newblock Wavelet-based autofocusing and unsupervised segmentation of microscopic images.
\newblock In {\em IEEE/RSJ International Conference on Intelligent Robots and Systems}, volume~3, pages 2143--2148.

\bibitem[Yang et~al., 2022]{maniqa}
Yang, S., Wu, T., Shi, S., Lao, S., Gong, Y., Cao, M., Wang, J., and Yang, Y. (2022).
\newblock {MANIQA}: Multi-dimension attention network for no-reference image quality assessment.
\newblock In {\em IEEE Conference on Computer Vision and Pattern Recognition}, pages 1191--1200.

\bibitem[Yang et~al., 2020]{yang2020advancing}
Yang, W., Yuan, Y., Ren, W., Liu, J., Scheirer, W.~J., Wang, Z., Zhang, T., Zhong, Q., Xie, D., Pu, S., et~al. (2020).
\newblock Advancing image understanding in poor visibility environments: A collective benchmark study.
\newblock {\em IEEE Transactions on Image Processing}, 29:5737--5752.

\bibitem[Ying et~al., 2020]{ying2019patches}
Ying, Z., Niu, H., Gupta, P., Mahajan, D., Ghadiyaram, D., and Bovik, A. (2020).
\newblock {From patches to pictures (PaQ-2-PiQ): Mapping the perceptual space of picture quality}.
\newblock In {\em IEEE Conference on Computer Vision and Pattern Recognition}, pages 3575--3585.

\bibitem[Zhai and Min, 2020]{percep}
Zhai, G. and Min, X. (2020).
\newblock Perceptual image quality assessment: a survey.
\newblock {\em Science China Information Sciences}, 63:1--52.

\bibitem[Zuo et~al., 2022]{zuo2022idea}
Zuo, Z., Chen, X., Xu, H., Li, J., Liao, W., Yang, Z.-X., and Wang, S. (2022).
\newblock Idea-net: Adaptive dual self-attention network for single image denoising.
\newblock In {\em IEEE Winter Conference on Applications of Computer Vision}, pages 739--748.

\end{thebibliography}

%\section*{\uppercase{Appendix}}

%If any, the appendix should appear directly after the
%references without numbering, and not on a new page. To do so please use the following command:
%\textit{$\backslash$section*\{APPENDIX\}}

\end{document}